\documentclass[11pt]{article}

\usepackage{ACL2023}
\usepackage{placeins}

\usepackage{times}
\usepackage{latexsym}
\usepackage{tabularx}
\usepackage{tcolorbox}
\usepackage[T1]{fontenc}

\usepackage{mdframed}
\usepackage[utf8]{inputenc}
\usepackage{euro}

\usepackage{microtype}

\usepackage{inconsolata}

\usepackage{xcolor}
\usepackage[textsize=footnotesize]{todonotes}
\usepackage{url}
\usepackage{graphicx}

\newenvironment{itemizesquish}[2]{\begin{list}{\labelitemi}{\setlength{\itemsep}{#1}\setlength{\labelwidth}{#2}\setlength{\leftmargin}{\labelwidth}\addtolength{\leftmargin}{\labelsep}}}{\end{list}}

\definecolor{towelone}{HTML}{fdf498}
\definecolor{toweltwo}{HTML}{f37736}
\definecolor{towelthree}{HTML}{ee4035}
\definecolor{towelfour}{HTML}{7bc043}
\definecolor{towelfive}{HTML}{0392cf}

\definecolor{redstrings}{rgb}{0.64,0.08,0.08}
\definecolor{bluekeywords}{rgb}{0.13,0.13,1}
\definecolor{greencomments}{rgb}{0.25,0.5,0.37}
\usepackage{listings}
\lstdefinelanguage{json}{
    basicstyle=\small\ttfamily,
    stepnumber=1,
    numbersep=8pt,
    showstringspaces=false,
    stringstyle=\color{redstrings},
    keywordstyle=\color{bluekeywords},
    commentstyle=\color{greencomments},
    morestring=[b]",
    morecomment=[s]{/*}{*/},
    morecomment=[l]//,
}

\title{Knowledge Base Question Answering for Space Debris Queries}

\author{Paul Darm$^{1}$ \\ \texttt{paul.darm@strath.ac.uk}
\And
Antonio Valerio Miceli-Barone$^2$ $\qquad$ \\ \texttt{amiceli@ec.ac.uk}
\AND
Shay B. Cohen$^2$ \\ \texttt{scohen@inf.ed.ac.uk}
\And
Annalisa Riccardi$^1$ \\ \texttt{annalisa.riccardi@strath.ac.uk}
\AND
\textnormal{\normalsize $^1$ Mechanical and Aerospace Engineering, University of Strathclyde}
\\
\normalsize $^2$ School of Informatics, University of Edinburgh
University
}

\begin{document}

\maketitle
\begin{abstract}
Space agencies execute complex satellite operations that need to be supported by the technical knowledge contained in their extensive information systems.
Knowledge bases (KB) are an effective way of storing and accessing such information at scale. In this work we present a system, developed for the European Space Agency (ESA), that can answer complex natural language queries, to support engineers in accessing the information contained in a KB that models the orbital space debris environment. Our system is based on a pipeline which first generates a sequence of basic database operations, called a 
 sketch, from a natural language question, then specializes the sketch into a concrete query program with mentions of entities, attributes and relations, and finally executes the program against the database. This pipeline decomposition approach enables us to train the system by leveraging out-of-domain data and semi-synthetic data generated by GPT-3, thus reducing overfitting and shortcut learning even with limited amount of in-domain training data. Our code can be found at \url{https://github.com/PaulDrm/DISCOSQA}.
\end{abstract}

\section{Introduction}
\label{SEC:INTRO}

Space debris are uncontrolled artificial objects in space that are left in orbit during either normal operations or due to malfunctions.
Collisions involving space debris can generate secondary debris which can cause more collisions, potentially leading to a runaway effect known as “Kessler Syndrome” \citep{Kessler1978, kessler2010kessler}, which in the worst-case scenario could make large ranges of orbits unusable for space operations for multiple generations.

\FloatBarrier
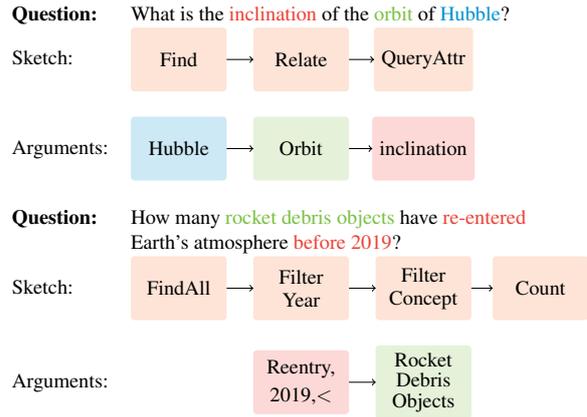
\begin{figure}[ht]
    \begin{center}

\scalebox{0.7}{

\begin{tabular}{ll}

\textbf{Question:} & What is the \textcolor{towelthree}{inclination} of the \textcolor{towelfour}{orbit} of \textcolor{towelfive}{Hubble}? \\

Sketch: & $\vcenter{\hbox{\begin{tikzpicture}[cc/.style={minimum height=1.21cm,fill=toweltwo!20,rounded corners=2pt,thick,inner sep=4,outer sep=0,minimum width=18mm},cred/.style={align=center,fill=towelthree!20,rounded corners=2pt,thick,inner sep=4,outer sep=0,minimum width=18mm},cgreen/.style={minimum height=1.21cm,fill=towelfour!20,rounded corners=2pt,thick,inner sep=4,outer sep=0,minimum width=18mm},corange/.style={minimum height=1.21cm,fill=towelfive!20,rounded corners=2pt,thick,inner sep=4,outer sep=0,minimum width=18mm}]

\path  (0,0) node[cc] (A) {Find} (2.3,0) node[cc] (B) {Relate} (4.6,0) node[cc] (C) {QueryAttr};
\draw[->] (A)--(B);
\draw[->] (B)--(C);

\end{tikzpicture}}}$ \\
\\

Arguments: & $\vcenter{\hbox{\begin{tikzpicture}[cc/.style={minimum height=1.21cm,fill=toweltwo!20,rounded corners=2pt,thick,inner sep=4,outer sep=0,minimum width=18mm},cred/.style={minimum height=1.21cm,align=center,fill=towelthree!20,rounded corners=2pt,thick,inner sep=4,outer sep=0,minimum width=18mm},cgreen/.style={minimum height=1.21cm,fill=towelfour!20,rounded corners=2pt,thick,inner sep=4,outer sep=0,minimum width=18mm},corange/.style={minimum height=1.21cm,fill=towelfive!20,rounded corners=2pt,thick,inner sep=4,outer sep=0,minimum width=18mm}]

\path  (0,0.) node[corange] (A) {Hubble} (2.3,0) node[cgreen] (B) {Orbit} (4.6,0) node[cred] (C) {inclination};
\draw[->] (A)--(B);
\draw[->] (B)--(C);

\end{tikzpicture}}}$

\\

\\

\textbf{Question:} & How many \textcolor{towelfour}{rocket debris objects} have \textcolor{towelthree}{re-entered}\\
&   Earth's atmosphere \textcolor{towelthree}{before 2019}? \\

Sketch: & $\vcenter{\hbox{\begin{tikzpicture}[cc/.style={minimum height=1.21cm,align=center,fill=toweltwo!20,rounded corners=2pt,thick,inner sep=4,outer sep=0,minimum width=18mm},cred/.style={minimum height=1.21cm,align=center,fill=towelthree!20,rounded corners=2pt,thick,inner sep=4,outer sep=0,minimum width=18mm},cgreen/.style={minimum height=1.21cm,align=center,fill=towelfour!20,rounded corners=2pt,thick,inner sep=4,outer sep=0,minimum width=18mm},corange/.style={minimum height=1.21cm,align=center,fill=towelfive!20,rounded corners=2pt,thick,inner sep=4,outer sep=0,minimum width=18mm}]

\path (0,0) node[cc] (A) {FindAll} (2.3,0) node[cc] (B) {Filter\\[1ex] Year} (4.6,0) node[cc] (C) {Filter\\[1ex]Concept} (6.8,0) node[cc] (D) {Count};
\draw[->] (A)--(B);
\draw[->] (B)--(C);
\draw[->] (C)--(D);

\end{tikzpicture}}}$ \\

\\
Arguments: &

$\vcenter{\hbox{\begin{tikzpicture}[cc/.style={minimum height=1.21cm,fill=toweltwo!20,rounded corners=2pt,thick,inner sep=4,outer sep=0,minimum width=18mm},cred/.style={minimum height=1.21cm,align=center,fill=towelthree!20,rounded corners=2pt,thick,inner sep=4,outer sep=0,minimum width=18mm},cgreen/.style={minimum height=1.21cm,align=center,fill=towelfour!20,rounded corners=2pt,thick,inner sep=4,outer sep=0,minimum width=18mm},corange/.style={minimum height=1.21cm,align=center,fill=towelfive!20,rounded corners=2pt,thick,inner sep=4,outer sep=0,minimum width=18mm},cwhite/.style={minimum height=1.21cm,align=center,fill=white,rounded corners=2pt,thick,inner sep=4,outer sep=0,minimum width=18mm}]

\path (0,0) node[cwhite] (ZZ) {} (2.3,0) node[cred] (AA) {Reentry,\\[1ex]2019,$<$} (4.6,0) node[cgreen] (BB) {Rocket\\[0.7ex]Debris\\[0.7ex]Objects};
\draw[->] (AA)--(BB);

\end{tikzpicture}}}$

\end{tabular}
}
    \end{center}
    \caption{Two representative queries for DISCOS and their decomposition according to the Program Induction method.}
    \label{fig:ioa_exampl_queries}
\end{figure}

Therefore, space agencies 
have established departments responsible for cataloging the space debris environment, which can be used for space traffic management, collision avoidance, re-entry analysis, and raising public awareness of the problem.\footnote{\url{https://tinyurl.com/44tc24d4}}

The European Space Agency (ESA) has catalogued over 40,000 trackable and unidentified objects in its DISCOS (Database and Information System Characterizing Objects in Space) Knowledge Base (KB) \citep{Klinkrad1991discos, flohrer2013discos}. 
Accessing this information efficiently often requires technical expertise in query languages and familiarity with the specific schema of DISCOS, which may fall outside the skillset of the engineers searching for relevant information in the database.
In this project, we developed a question answering 
system for the DISCOS KB. This deployed prototype enables ESA engineers to query the database with complex natural language (English)
questions, improving their ability to make informed decisions regarding space debris.

Recent breakthroughs in open question answering have been achieved using large language models that have been fine-tuned as dialog assistants, such as ChatGPT.\footnote{\url{https://chat.openai.com}}
These models, however, are black boxes that store knowledge implicitly in their parameters which makes it hard to guarantee that their answers are supported by explicit evidence, understand their failures and update them when the supporting facts change.
In contrast, parsing a question into a query program and then executing it on an explicit KB is guaranteed to provide a factual correct answer provided the KB and query program are correct. 
Our approach is particularly useful for applications such as satellite operations where accuracy and reliability are critical.

The main challenge for this project was that no training set or example questions were available for the DISCOS KB.
This issue, combined with the large amount of unique and diverse objects in the database, precluded a straightforward application of common supervised learning techniques.
 Although possible strategies for solving this task, such as direct semantic parsing of the query with seq2seq models, were identified in the literature, they suffer from problems with compositional generalization \cite{DBLP:journals/corr/abs-2104-07478, https://doi.org/10.48550/arxiv.2007.08970}.
Furthermore, very little work has been done on generalizing to KB element components that were never seen during training \cite{cao-etal-2022-program, das-etal-2021-case, huang-etal-2021-unseen-entity}.
 


To overcome these challenges, we apply and adapt a methodology from the literature called Program Transfer \citep{cao-etal-2022-program} to significantly reduce the required dataset for adequate generalization over the complete DISCOS KB.
This is a two-step approach. For each user query first a program sketch is predicted, consisting of a sequence of query functions where the arguments are either variables or placeholders, 
then the representation of the query is compared to the representations of the KB entities, in order to fill out the placeholders with arguments relevant to the query text. The underlying query language of this approach is called Knowledge-oriented-Programming-Language (KoPL) for which two representative example questions are shown together with their decomposition into sketch and arguments in Figure~\ref{fig:ioa_exampl_queries}. 

We also conduct a data collection study with domain experts, and we apply a data augmentation pipeline leveraging the underlying ontology of the KB and prompting a Large Language Model (LLM) to generate automatically more training examples. The architecture was retrained with different domain-specific LMs and baselines to determine the benefits of using a domain-specific pre-trained encoder. 

The main contributions of this paper are: 
\begin{itemizesquish}{-0.3em}{0.5em}
    \item Applying and adapting a methodology described in the literature for complex knowledge base question answering (CKBQA) on a novel industry-relevant database, with a large and dynamic set of unique entities;
    \item Collecting a new dataset on this database from domain-experts and leveraging the in-context learning capability of LLMs for data augmentation on it;
    \item Evaluating the use of domain-specific LMs as different encoders on our curated dataset;
    \item Demonstrating the effectiveness of the approach by achieving comparable results to general-purpose LLMs
\end{itemizesquish}

\section{Related Work}

\paragraph{Low-resource CKBQA}
Pre-trained language models have demonstrated state-of-the-art performance in semantic parsing for complex question answering on KBs where the same logic compounds are contained in both the training and validation sets \cite{https://doi.org/10.48550/arxiv.2007.08970}. However, they struggle with compositional generalization, where the ``compounds'' (combinations) of components are diverse between training and validation, even if all components (entity, relation, program filters) have been seen during training \citep{DBLP:journals/corr/abs-2104-07478}. 
\citet{https://doi.org/10.48550/arxiv.2104.08762} explored retrieval-based methods to pick the top $n$ similar examples from the training set and use them as additional input for the prediction. In theory, this would make it possible to reason over changes on the KB by only adding new examples to the training set without the need of retraining the whole model. 
Another approach is adapting the architecture of language models to incorporate 
the structure of a KB directly for the prediction. For example, \newcite{huang-etal-2021-unseen-entity} ranked FreeBase KB entities by using an EleasticSearch search API to identify these entities. 
When generating the query program, instead of entities a special token is predicted, which in the post-processing step get replaced by the top ranked entity identified by ElasticSearch. 
Although, achieving good results, it is unclear how this would translate to queries with multiple entities and also has the typical limitations of ElasticSearch. 
Another method is the Program Induction and Program Transfer method, where a sequence of \emph{functions}, or a \emph{sketch}, is generated from the input query. A single function here stands for a basic logic operation on the KB. The premise is that the sketch is mostly dependent on the formulation of the input query and less dependent on the KB, therefore the training on a source domain can transfer to inference on a target domain. In a second step, the particular inputs that each function in the sketch receives are identified from the elements of the KB through comparing their representations with the one of the model at the specific function. During training, the goal is to create sophisticated representations for the components of the KB as well as for the query that can generalize to components which were not seen during training \cite{cao-etal-2022-program}. 

\paragraph{Domain-specific Language Models}
Using self-supervised pre-trained language models is the de-facto standard approach in modern natural language processing (NLP). These models are trained on large volumes of text, learning representations that can generalize over natural language variations and capture long-term dependencies between the input tokens. During fine-tuning, these learned features and representations commonly lead to improved results on the downstream task. While the interplay between the amount of in-domain data, model capacity and training regime is complex \cite{zhao2022understanding}, as a general rule, training these models on in-domain task-related text improves the performance of this task \citep{maheshwari-etal-2021-scibert, spacetrans_berquand, arnold-etal-2022-extraction, https://doi.org/10.48550/arxiv.2301.13779}.
An alternative approach involves modifying the pre-training objective according to the domain. In the context of question-answering with tabular data, it was explored how a language model could function during pre-training as a SQL-query excecutor, predicting the results of an automated created SQL query on the corresponding concatenated table, to elicit an understanding of the underlining dependencies in tables. This approach resulted in improved performance on related downstream tasks \citep{https://doi.org/10.48550/arxiv.2107.07653}.\
For KBs, together with the standard Masked-Language-Modelling loss an architecture called Kepler was tested that also minimizes a contrastive loss on related KB triples, where both the correct triples and randomly perturbed incorrect triples are scored, and the loss penalizes scoring the perturbed triples higher than the correct ones \citep{wang2021KEPLER}.


\section{Dataset Collection and Use} 

We used two sources of data for the study. The first one is the original dataset which first introduced the KoPL-format for general-domain knowledge-base question answering (described in \S\ref{section:kqa}). The second one is for our particular domain with information about objects in space (\S\ref{section:discos}).
We also describe in \S\ref{section:collection} how we further collected training data for fine-tuning a language model, and how we augmented this dataset with new question-answer pairs (\S\ref{section:aug}).

\subsection{The KQA Pro Dataset}
\label{section:kqa}

The KQA pro dataset \cite{cao-etal-2022-kqa} is a large scale dataset for complex question answering over a knowledge base. It contains over 120,000 diverse questions for an subset of entities from the Freebase KB and their associated relations, attributes and concepts. 
The reasoning process to arrive at a solution is provided in the form of a Knowledge-oriented-Programming-Language (KoPL), which was designed specifically for this dataset. 
The question-program pairs were automatically generated by randomly sampling the extracted KB and using novel compositional templates to create a canonical form of the question and associated answer. To increase ambiguity these questions were then paraphrased and controlled by Amazon Mechanical Turk workers \cite{cao-etal-2022-kqa}.

\subsection{The DISCOS Database} 
\label{section:discos}

The ESA Database and Information System Characterising Objects in Space (DISCOS)\footnote{\url{https://discosweb.esoc.esa.int/}} is a regularly updated source for information about launches, launch vehicles, objects, spacecraft registration numbers, mission-specific information (mass, mission objective, operator), and most importantly orbital and fragmentation data histories for all trackable as well as unclassified objects. With currently over 40,000 objects  being tracked, this tracking provides rich information for ESA offices monitoring and managing space debris, collision avoidance, re-entry analyses, and contingency support. 
Other actors, such as research institutes, government organisations, or industrial companies from ESA Member states can apply for an account to access the information provided by DISCOS free of charge. A comparison between the the DISCOS KB and the KB used for the KQA Pro dataset can be seen in Table~\ref{table:entity_relation_concept_counts}

\begin{table}[ht]
\centering
\resizebox{\columnwidth}{!}{\begin{tabular}{lrrr}
\hline
\textbf{Dataset} & \textbf{\#Entities} & \textbf{\#Relations} & \textbf{\#Concepts} \\ \hline
KQA Pro & 16,960 & 363 & 794 \\ \hline
DISCOS & 73,354 & 32 & 39 \\ \hline
\end{tabular}}
\caption{Entity, relation, and concept counts for the KQA Pro and DISCOS KBs}
\label{table:entity_relation_concept_counts}
\end{table}

\subsection{Data Collection}
\label{section:collection}

As there was no question-program-answer training set available on the DISCOS KB, potential queries had to be collected from domain-experts via a simple user interface. The interface allowed domain experts to input queries of interest, along with their username and feedback, see Appendix~\ref{appendix:interface}. 
Based on the domain experts' feedback, a manually labeled baseline dataset of around 102 question-KoPL program pairs was created.
\par

\subsection{Data Augmentation}
\label{section:aug}


To extend the limited baseline dataset and increase diversity of potential queries, we augmented the dataset by creating paraphrases of the questions, shown to add robustness to question answering systems \cite{fader2013paraphrase,narayan2016paraphrase}. 
For each unique program sketch, 
the schema of the ontology was used to alter the arguments of the single functions in that sketch. 
For example, for the program \textit{Find(`Saturn V') $\rightarrow$ QueryAttr(`mass')}, the concept of the \textit{Find} function argument (\textit{Saturn V}) was identified as \textit{LaunchVehicle}. Subsequently, the ontology was queried for other entities of the same concepts and  also for associated attributes to substitute the argument for the \textit{QueryAttr} function. 
\par

In order to generate appropriate questions for each ``augmented'' program, we used the few-shot and in-context learning capabilities of GPT-3 \citep{Brown2020GPT3}.
A prompt was curated, consisting of question-program pairs from the manually labeled dataset, so that GPT-3 could generate a question for an unlabelled augmented program. 
The sampling included all programs from the manual dataset with the same sketch as the augmented one, as well as using examples with the same relation type, which showed great abilities to generate a correct question. Additionally, an instruction section was added to the prompt, consisting mainly of a list of expansions for acronyms commonly used in the KB's ontology. This ensured that acronyms were expanded correctly and reduced the likelihood of hallucinations by the LLM for these acronyms. Examples of generated questions with their corresponding programs can be found in Appendix~\ref{appendix:exam_aug}. The prompt schema can be found in Appendix~\ref{appendix:prompt}.




The benefits of an automatically generated dataset include cost-effective data sample generation, while also ensuring a balanced distribution of complex and simple queries as well as common (ex: Saturn V, Hubble Space Telescope) and uncommon arguments (ex: L-186, PSLV-Q). 
With the use of LLMs, the generated questions also have already slightly different semantics and syntactic structure as the generation process is inherently statistical and can be adjusted with parameters such as the temperature. LLMs can also leverage their stored world knowledge, e.g. we observed the entity name "\includegraphics[trim={0 1.2eM 0 0}, clip, height=1.1eM]{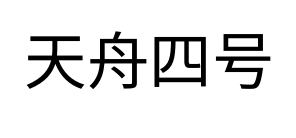}" to be automatically translated into the english "Tianzhou 4" designation. More examples can be seen in Appendix \ref{appendix:exam_aug}. However, it is important to note that the question generation is not foolproof and could be further optimized e.g. through  additional prompt engineering or a subsequent data cleaning procedure. \par

\section{Methodology and Model Architecture}

We describe in \S\ref{section:problem} the problem that
our model aims to solve. In \S\ref{section:pt-enc}, we describe
what modification we applied to the methodology from \cite{cao-etal-2022-program}.

\subsection{Problem Definition}
\label{section:problem}

The task is defined in the following way, given a natural language question $Q$ we want to predict a program $y$ that traverses the knowledge base $K$ and produces an answer $A$ for $Q$. This means: $$A=y(Q,K), \quad K = \{E,R,C,A\},$$ where ${E,R,C,A}$ represent respectively the mutually disjoint sets of entities, relations, concepts and attributes in $K$. More specifically for a set entities in the training set $E_t$, the task is to be able to generalise to the set of $E_v$, which were unseen during training, with $E=E_t \cup E_v$, $E_t \cap E_v = \emptyset$. Therefore, we apply a program induction and transfer methodology, predicting for $Q$ a program, a tuple of actions, $$y (Q,K)= (o^{1}(arg^{1}),\dots, o^{t}(arg^{t})),$$  $$o^{i}\in O, \forall i=1,\ldots,t; \quad arg^{t} \in E \cup R \cup C \cup A,$$ where $O$ is a set predefined basic functions executable on the KB, with each of them taking one disjoint set as inputs from the pool of arguments from the KB. 
In the first step, the sketch $[o^{1},\dots, o^{t}] \in O^t$ is generated by encoding the question $Q$ with a pre-trained language model and using its representation as the starting point for a GRU-based decoder with attention mechanism \citep{bahdanau2014neural, cao-etal-2022-program}.

The input arguments for each function, $o^1,\ldots,o^t$, are chosen in a second step by calculating the probability between the encoded representations of the $i$-th candidate $R_i^{t}$ in the KB at position $t$ in the sequence, and the representation of the decoder at position $t$, $h^{t}$. The probability is computed as: $$ p(arg^{t}|R_i^t,h^{t}) = \textrm{softmax}(R^{t}_i,h^{t}),$$ and is followed by choosing the most likely candidate as the input argument for the function.

We refer the reader to \citet{cao-etal-2022-program} for the details.
In addition to the standard BERT model 
we also train RoBERTA-based \citep{liu2019RoBERTa} domain-specific encoders (see Appendix~\ref{appendix:lms}).

\subsection{Enhancements to Program Transfer}
\label{section:pt-enc}


At the beginning, the prediction for entities, relations, and operations ($\{ <,>,= \}$) had to be re-implemented as it was not available in the associated source code.\footnote{\url{https://github.com/thu-keg/programtransfer}}

The standard methodology of comparing the representations to all inputs during training posed another challenge for entities, as during the training step the gradients for all them would need to be stored, which were exceeding the available virtual random access memory on the system. 

As a result, instead of comparing the representation at each sequence position to all entities in the KB, only the subset of entities in the current batch are compared with each other.\footnote{Note that these entities are part of the training set entities, hence are all known apriori.} 
In addition, random samples from the complete entity set where selected and added to the batch to also give signals to entities not occurring in the training set. 

More formally, let $X^{(j)}$ be the $j$th batch from the training set, $E$ be the overall set of entities, and $X^{(j)}_E$ be the set of entities of the training batch.
At each training batch $j$, we use the set of entities $e^{(j)}$ to compare the probability of their representations against the candidate input argument in batch $j$, as defined by:

\begin{equation}
e^{(j)} = X^{(j)}_E \cup f^{(j)},
\end{equation}
 
\noindent where $f^{(j)} = \{ e_{1}, e_{2}, \dots, e_{n} \}$ are $n$ randomly sampled entities from $E$ without replacement. Furthermore, for extracting time values from queries we use the SUTime library \cite{chang-manning-2012-sutime}, whereas for extracting numerical patterns we use a simple regex schema. In addition, for parsing the predicted inputs into a form than can be read by the KoPL engine a parser function had to be written as well as a method for assigning the dependencies between the single basic functions. Other small adaptation of the original approach included the addition of a normalization layer between the linear layers for the prediction of the single arguments as well as masking-out padding tokens from the loss calculation of the function generation component. 
A schematic overview of the pipeline can be seen in Appendix~\ref{appendix:Pipeline_components}.

\section{Experiments}
\label{section:experiments}

We next describe our dataset (\S\ref{section:dwq}), the way we trained our models (\S\ref{section:model-training}) and our results (\S\ref{section:results}).

\subsection{The DISCOS-Questions-Programs Dataset}
\label{section:dwq}

The creation of the training and validation datasets involved the following steps: First, all the unique sequences of functions were extracted from the manually labeled dataset. Then, ten augmented programs were generated for each unique program in the manually labeled dataset by substituting the inputs and generating questions as described in the data augmentation methodology. It is noteworthy that the entities included in the manually labeled dataset were not considered as candidates for the data augmentation. The questions for the augmented examples were generated using the large language model \textit{code-davinci-002}.
\footnote{\url{https://platform.openai.com/playground}} Through trial and error, a temperature of 0.75 was used for generation as it produced diverse examples while still capturing the meaning of the associated KoPL-program. Finally, the augmented dataset was split into training and validation datasets, where the original manually labeled dataset and 0.05\% of randomly sampled examples from the augmented dataset were used as the validation set and the rest of the augmented dataset was used for training. In a subsequent filtering step, programs that appeared in the validation set were filtered from the training dataset. The resulting DISCOS-Questions-Programs (DQP) dataset consists of 905 samples for training and 151 samples for validation. \par

\subsection{Model Training}
\label{section:model-training}

The preliminary results indicated that training directly on the DQP dataset did not lead to convergence in entity prediction. We then proceeded to first train on the out-of-domain original KQA Pro dataset and then further train on the DQP dataset. The hyperparameters for pretraining on the KQA pro dataset were adopted from the original paper. For training the DQP dataset, the hyperparameters were left unchanged, except for an adapted learning rate for the decoder, which was set to $10^{-4}$ instead of $10^{-3}$. For the experiments, different domain-adapted models were used as the encoder and then compared to each other as well as baseline models. For more information about the domain-adapted models see Appendix~\ref{appendix:lms}. 

\subsection{Results}
\label{section:results}

The analysis of different models was challenging as multiple components (functions, entities, relations, etc.) need to be predicted to arrive at the full program that can be run against the KB.
Therefore, the analysis was divided into two parts. Firstly, the accuracy at the lowest validation loss for each component was compared separately between all the different trained models, providing an overview of each model's best predictive performance for each component. The results can be seen in Table \ref{tab:acc_low_val}. Although no model consistently outperformed the others on all components, CosmicRoBERTa \cite{spacetrans_berquand} achieved the highest performance on four out of six accuracies. \par

The validation loss curves showed that during training, the validation accuracy can drop for one component while it rises for another. 
The validation loss curves can be found in Appendix \ref{appendix:losscurves}.\par
For deployment of a single model, it is necessary to identify a checkpoint where the model predicts accurately across all components. To obtain a more holistic view of the performance, the models were also compared by summing up the normalized validation losses for each component.

\begin{table}[ht]
\centering
\setlength{\tabcolsep}{3pt}
\begin{tabular}{lrrrr}
\hline
\textbf{Accuracy} & \textbf{BERT} & \textbf{Kepler} & \textbf{RoBERTa} & \textbf{CR} \\ \hline
Function  & 0.797 & \textit{0.812} & 0.796 & \textbf{0.826} \\
Entity & 0.874 & 0.887 & \textit{0.907} & \textbf{0.927} \\
Attribute  & \textbf{0.955} & \textit{0.948} & \textit{0.948} & \textit{0.948} \\
Relation  & 0.938 & \textit{0.983} & \textit{0.983} & \textbf{1} \\
Concept  & 0.872 & \textit{0.896} & \textbf{0.92} & 0.89  \\ 
Operations & \textbf{1} & \textbf{1} & \textbf{1} & \textbf{1} \\
\hline
\end{tabular}
\caption{{Accuracy at lowest validation loss for each respective component. CR stands for CosmicRoBERTa. The best score for each component is highlighted in \textbf{bold}, second best in \textit{italics}.}}
\label{tab:acc_low_val}
\end{table} 

The results of summing up the normalized validation losses with equal weights are shown in Table \ref{tab:acc_mean_val}. On average, the RoBERTa-base model has the lowest validation loss, followed by the Kepler model. However, no model consistently outperforms the others in terms of prediction accuracy. From an application perspective, the most important metrics are the accuracy in predicting functions and entities. To obtain the correct answer, the most crucial step is to predict the correct sequence of functions, and for multi-hop queries, it is essential to identify the correct starting entity.
In addition, the number of unique entities is magnitudes higher than the number of unique attributes or relations in the KB, which makes identifying the right entity more difficult. Among the models considered, CosmicRoBERTa stands out as the model that performs well both in predicting functions and entities. 
Unfortunately, the Kepler model only sporadically showed improvements over RoBERTa-base. The reason for this could be the very limited pre-training corpus, which as a result was significantly smaller than the one from CosmicRoBERTa As a result, for the purpose of deploying a single model, the decision was made to choose CosmicRoBERTa.

\begin{table}[ht]
\centering
\setlength{\tabcolsep}{3pt}
\resizebox{\columnwidth}{!}{\begin{tabular}{lrrrr}
\hline
\textbf{Accuracy} & \textbf{BERT} & \textbf{Kepler} & \textbf{RoBERTa} & \textbf{CR} \\ \hline
Min. sum valid loss & 0.21 & \textit{0.171} & \textbf{0.11} & 0.252 \\
\hline
Function  & \textbf{0.79} & 0.734 & 0.775 & \textit{0.789} \\
Entity  & 0.874 & 0.887 & \textit{0.894} & \textbf{0.927} \\
Attribute  & 0.935 & \textbf{0.948} & \textit{0.941} & 0.915 \\
Relation  & \textbf{0.978} & 0.95 & \textit{0.969} & \textit{0.969} \\
Concept  & 0.853 & 0.8841 & \textit{0.908} & \textbf{0.927} \\
Operations  & \textbf{1} &	\textit{0.97} &	\textbf{1}	& 0.94 \\
\hline
\end{tabular}}
\caption{Accuracy at the lowest validation loss summed over all compoennts. The best score for each component is highlighted in \textbf{bold}, second best in \textit{italics}. CR stands for CosmicRoBERTa.}
\label{tab:acc_mean_val}
\end{table} 

The results of the prediction are overall very impressive, with high accuracy scores over all components. It is especially worth highlighting that from over 40,000 entities in the database, only 400 appear in the training and validation set and only 1 of 71 entities in the validation set also appear in the training set. Despite this, some models achieve an accuracy of over 90\% in predicting entities, demonstrating their strong ability to generalize to entities not seen during training, which was a critical user requirement for our system. 

In addition, we benchmarked our method to recently released general purpose models such as ChatGPT\footnote{\url{https://openai.com/blog/chatgpt}} and GPT-4 \cite{openai2023gpt4}. For each model, training set examples were randomly added to the prompt until the respective model's context limit is reached. Then the models were prompted to generate the right program for a question from the validation set. Our methodology has an overall accuracy of predicting the right program completely of 48\%, which is higher than the performance of around 25\% of ChatGPT and comparable with the performance of GPT-4 of around 50\%. A detailed comparison can be found in Table \ref{tab:comparison}. 
This further demonstrates the efficiency of our methodology as it can be run at a fraction of the necessary compute as well as locally on consumer-grade hardware.   


\begin{table}[h!]
\centering
\setlength{\tabcolsep}{3pt}
\resizebox{\columnwidth}{!}{\begin{tabular}{lccc}
\hline
\textbf{} & \textbf{CosmicRoBERTa} & \textbf{GPT-4} & \textbf{ChatGPT-3.5} \\
\hline
Functions & \textbf{0.79} & \textbf{0.79} & 0.516 \\
Entities & \textbf{0.93} & 0.86 & 0.41 \\
Relations & \textbf{0.97}& 0.87 & 0.32 \\
Concepts & \textbf{0.93} & 0.82 & 0.61 \\
Attributes & \textbf{0.92} & 0.84 & 0.72 \\
\hline
Overall & 0.48 & \textbf{0.5} & 0.25 \\
\hline
\end{tabular}}
\caption{Accuracy of deployed model (CosmicRoBERTa) versus general purpose models ChatGPT-3.5 and GPT-4. The best score for each component is highlighted in \textbf{bold}.}
\label{tab:comparison}
\end{table}

\section{Conclusion}

We developed a system for ESA to address the challenge of answering complex natural language questions on their DISCOS KB. The main obstacles included a lack of training data, the diversity and regular updates of the database entries, and the need for an economically feasible solution. 
The program transfer for complex KBQA methodology was selected for its potential to reduce the amount of required training samples through transfer learning and its capability to potentially generalize for examples which were never seen during training. 
A data collection study was conducted with domain experts, which was then used to augment the data through leveraging the underlying ontology of the KB and prompting a large language model to generate fitting questions. The architecture was retrained with different domain-specific models and baselines to determine the benefits of using a domain-specific pre-trained encoder. Although the results were mixed, the best performance was achieved by CosmicRoBERTa, a pre-trained model on a space domain corpus. 
With an accuracy of over 90\% of predicting the right entity on the validation set over the vast pool of candidate entities, 
the method demonstrates its strong ability to predict the correct input arguments for unseen examples. This is further demonstrated in the comparison with general-purpose models such as ChatGPT or GPT-4, where our method achieved competitive results. Therefore, this approach has the potential to be extended to other databases and query languages in the future, especially in scenarios where there are few to no training examples.

\section*{Limitations}

The study has several clear limitations. Firstly, the training and validation datasets used in this study are still relatively small. A larger dataset would give more robust results for comparing different encoders. Additionally, the experiments were only conducted on the ability to generalize to unseen entities and not on the ability to generalize to unseen sketch types, which is also of key importance when addressing low resource CKBQA. 

Moreover, the methodology used in this study relies on annotated question-program pairs, which are expensive to collect. Learning only from question-answer pairs or even a question with an indicated difficulty based on whether the model was able to answer the question, could be more easier to collect.
While the models achieve high accuracy on most of the knowledge base components, overfitting can occur at different stages during training, leading to high accuracy for one component at one training step but poor accuracy for another component at another step. In the future, revising the training procedure or the model setup may help address this issue.

\newpage
\section*{Ethics Statement}

In any safety-critical context like spacecraft operations, there is an inherent risk associated with the use of automatic methods supporting human operators. The transparency of the predicted programs could mitigate this issue as it allows even for an engineer with limited knowledge about the underlying query method to interpret the program to some degree. In any case, the developed systems might support human analysis and decision making by decreasing workload, but cannot replace it. As mentioned before the DISCOS KB can be accessed after creating a user account. We plan on publishing the created question-program paris and trained models online in accordance with ESA's guidelines. 


\section*{Acknowledgments}
This study was funded by the European Space Agency (ESA) under the Intelligent Operational Assistant project with contract No.AO/1-10776/21/D/SR.
This work was also supported by the UKRI Research Node on Trustworthy Autonomous Systems Governance and Regulation (grant EP/V026607/1) which provided additional funding for Antonio Valerio Miceli-Barone.
We thank Evridiki Ntagiou (ESA) and Paulo Leitao (Vision Space) for their roles as project leader and industrial partner, respectively. We are grateful for Callum Wilson's help (University of Strathclyde) in creating the GraphDB KG from the original DISCOSweb API, which was used to create the DISCOS KB. We also thank Sean Memery and Maria Luque Anguita (University of Edinburgh) for their contribution to the KEPLER data collection, and everyone who took part in the data collection for the first iteration of the DISCOSWeb-questions dataset. We also thank Marcio Fonseca and the reviewers for useful feedback on the paper.

\bibliography{anthology,custom}
\bibliographystyle{acl_natbib}

\appendix

\section{Instructions for Generating Questions}
\label{appendix:prompt}

\begin{tcolorbox}
\textit{"Here is a list of knowledge graph query programs in JSON format, each with its corresponding question in English.
Acronyms are expanded with the following dictionary: 
["GEO": "Geostationary Orbit", "IGO":"Inclined Geosynchronous Orbit", "EGO":"Extended Geostationary Orbit", 
"NSO":"Navigation Satellites Orbit", "GTO":"GEO Transfer Orbit", "MEO":"Medium Earth Orbit",
"GHO":"GEO-superGEO Crossing Orbits", "LEO":"Low Earth Orbit", "HAO":"High Altitude Earth Orbit", 
"MGO":"MEO-GEO Crossing Orbits", "HEO":"Highly Eccentric Earth Orbit", "LMO":"LEO-MEO Crossing Orbits",
"UFO":"Undefined Orbit","ESO":"Escape Orbits"]
Program: \{program\}
Question: \{question\}
(Repeated until maximum prompt limit is reached)
}
\end{tcolorbox}

\section{Domain-specific Language Models}
\label{appendix:lms}
\paragraph{CosmicRoBERTa}
A domain-specific language model from the SpaceTransformer family trained with basic masked-language-modelling on a corpus consisting of 75M words \cite{spacetrans_berquand}.\footnote{\url{https://huggingface.co/icelab/cosmicroberta}} 

\paragraph{Kepler}
We trained our own domain-specific language model by appending the KB pre-training objective as described by  \newcite{wang-etal-2021-kepler}. The in-domain text data for the Knowledge-augmented LM was specifically mined to be closely related to the topic of DISCOS. To achieve this, we collected scientific papers about space debris and documents from ESA about their internal mission operation procedures. Additionally, we added the online available dataset from SpaceTransformers\footnote{\url{https://pureportal.strath.ac.uk/en/datasets/dataset-of-space-systems-corpora-thesis-data}} to the training data.

The complete dataset comprises approximately 17.6 million words, with around 70\% used for training and the remainder for validation. To predict triples, we converted our database into a set of (\textit{head}, \textit{relation}, \textit{tail}) triples, where \textit{head} and \textit{tail} are entities represented by their English name or description, and \textit{relation} is a relation represented by a unique token for the relation type.

The KG triplet datasets consist of approximately 640,000 triples, which are split into 636,000 for training, 2,000 for validation, and 2,000 for testing. These triples represent approximately 59,000 entities and 32 relations. We trained our model using the available Kepler implementation.\footnote{\url{https://github.com/THU-KEG/KEPLER}}

\section{Pipeline Components}
\FloatBarrier
\label{appendix:Pipeline_components}
\begin{figure}[ht]
    \begin{center}
    \includegraphics[width=0.5\textwidth]{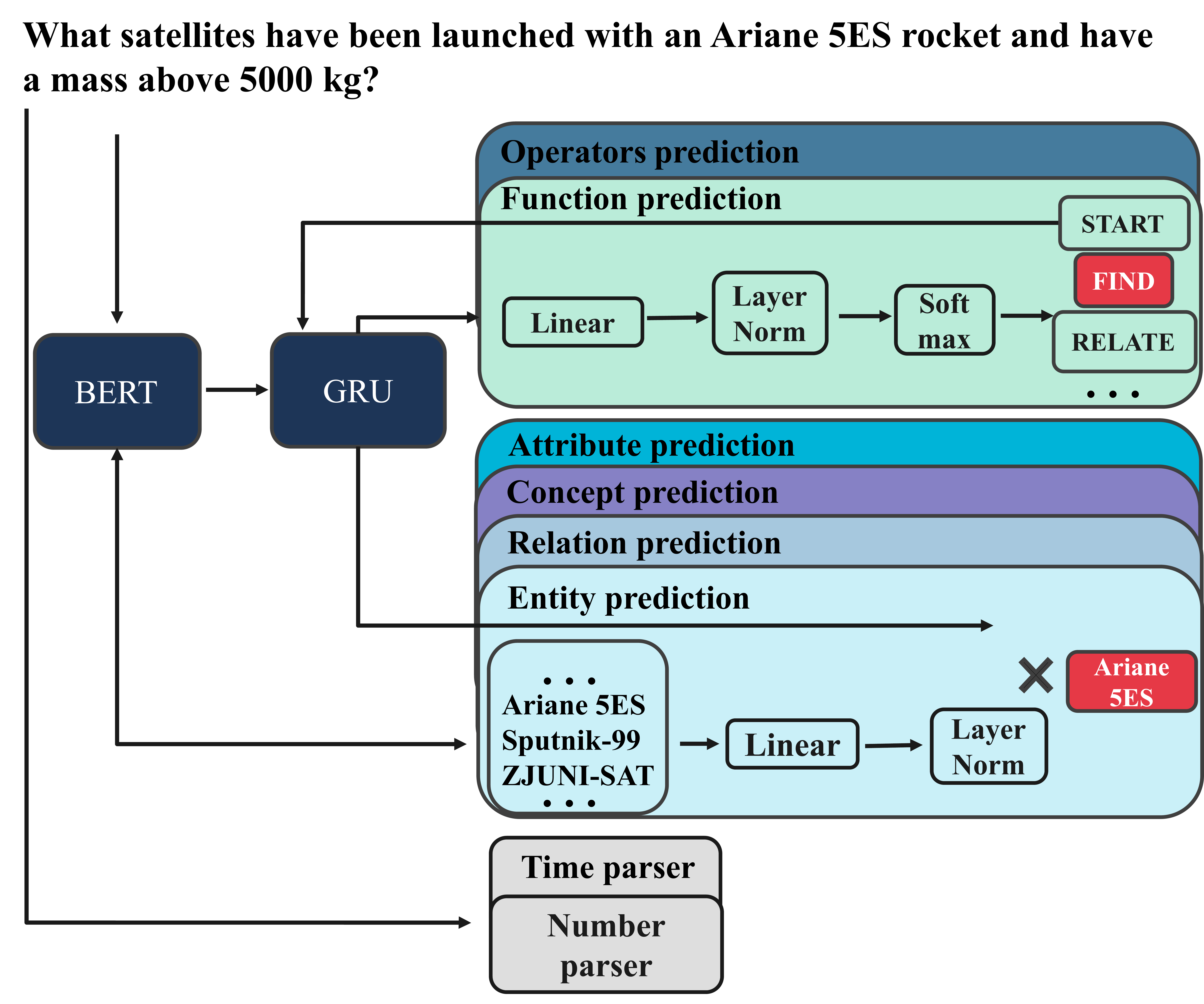}
    \end{center}
    \caption{Overview over components in query processing pipeline.} 
    \label{fig:ioa_components}
\end{figure}



\begin{onecolumn}
\section{Example Augmented Programs and Generated Questions}
\label{appendix:exam_aug}

\FloatBarrier

\begin{tcolorbox}[width=1.05\textwidth]
\begin{lstlisting}[language=json,firstnumber=1]
[{"function":"FindAll", "inputs":[], "dependencies":[]}, 
{"function":"FilterConcept", "inputs":["LMO"], "dependencies":[0]},
{"function":"Relate", "inputs":["orbit"], "dependencies":[1]}, 
{"function":"FilterNum", "inputs":["depth", "0.3", "="], "dependencies":[2]}, 
{"function":"FilterConcept", "inputs":["UnknownObjClass"], "dependencies":[3]}, 
{"function":"Count", "inputs":[], "dependencies":[4]}]
\end{lstlisting}
\textbf{Question}: How many objects are there that are currently in the LEO-MEO crossing orbits and are classified as unknown and have a depth of 0.3 m?
\end{tcolorbox}

\begin{tcolorbox}[width=1.05\textwidth]
\begin{lstlisting}[language=json,firstnumber=1]
{"function":"Find", "inputs":["State Remote Sensing Center"], "dependencies":[]},
{"function":"Relate", "inputs":["host_country"], "dependencies":[0]},
{"function":"FilterConcept", "inputs":["Entity"], "dependencies":[1]},
{"function":"What", "inputs":[], "dependencies":[2]}]
\end{lstlisting}
\textbf{Question}: "Which operators are based in the host country of the State Remote Sensing Center?"
\end{tcolorbox}

\begin{tcolorbox}[width=1.05\textwidth]
\begin{lstlisting}[language=json,firstnumber=1]
{"function":"FindAll", "inputs":[], "dependencies":[]},
{"function":"FilterDate", "inputs":["epoch", "2022-04-08", "="], "dependencies":[0]}, 
{"function": "FilterConcept", "inputs":["Launch"], "dependencies":[1]}, 
{"function": "Count", "inputs":[], "dependencies":[2]}]
\end{lstlisting}
\textbf{Question}: "How many launches are planned for 8th of April 2022?"\footnote{Interestingly the generated question implies that 8th of April 2022 lies in the future, which is in accordance with OpenAI's training set cut-off in September 2021}
\end{tcolorbox}


\FloatBarrier
\clearpage
\section{Interface for Data Collection and Access to System}
\label{appendix:interface}

For implementing a simple user interface, the popular Python library Streamlit was used. Besides providing the input query the user can also provide feedback, which potentially could be used to improve the model by identifying right or wrong answers and adding them to the training set. Another button allows the user to generate an answer for a question from the validation dataset to get a better feeling of what questions could be answered.

\FloatBarrier

\begin{figure*}[ht]
    \centering
    \includegraphics[page=1,width=0.9\textwidth]{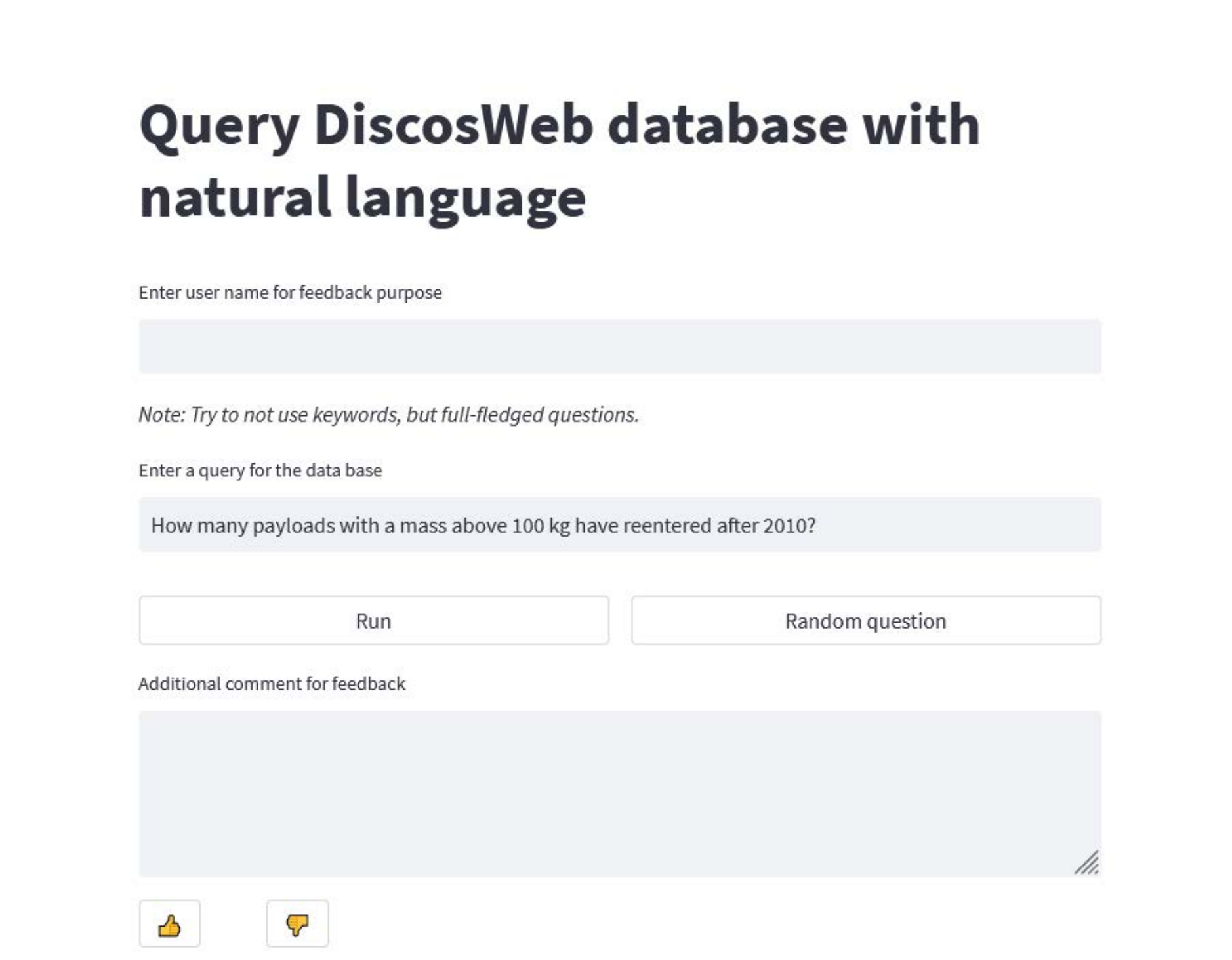}
    \label{fig:ioa_interface}
\end{figure*}

\clearpage
\section{Validation Loss Curves}
\label{appendix:losscurves}
\begin{figure}[ht]
    \begin{center}
    \includegraphics[width=\textwidth]{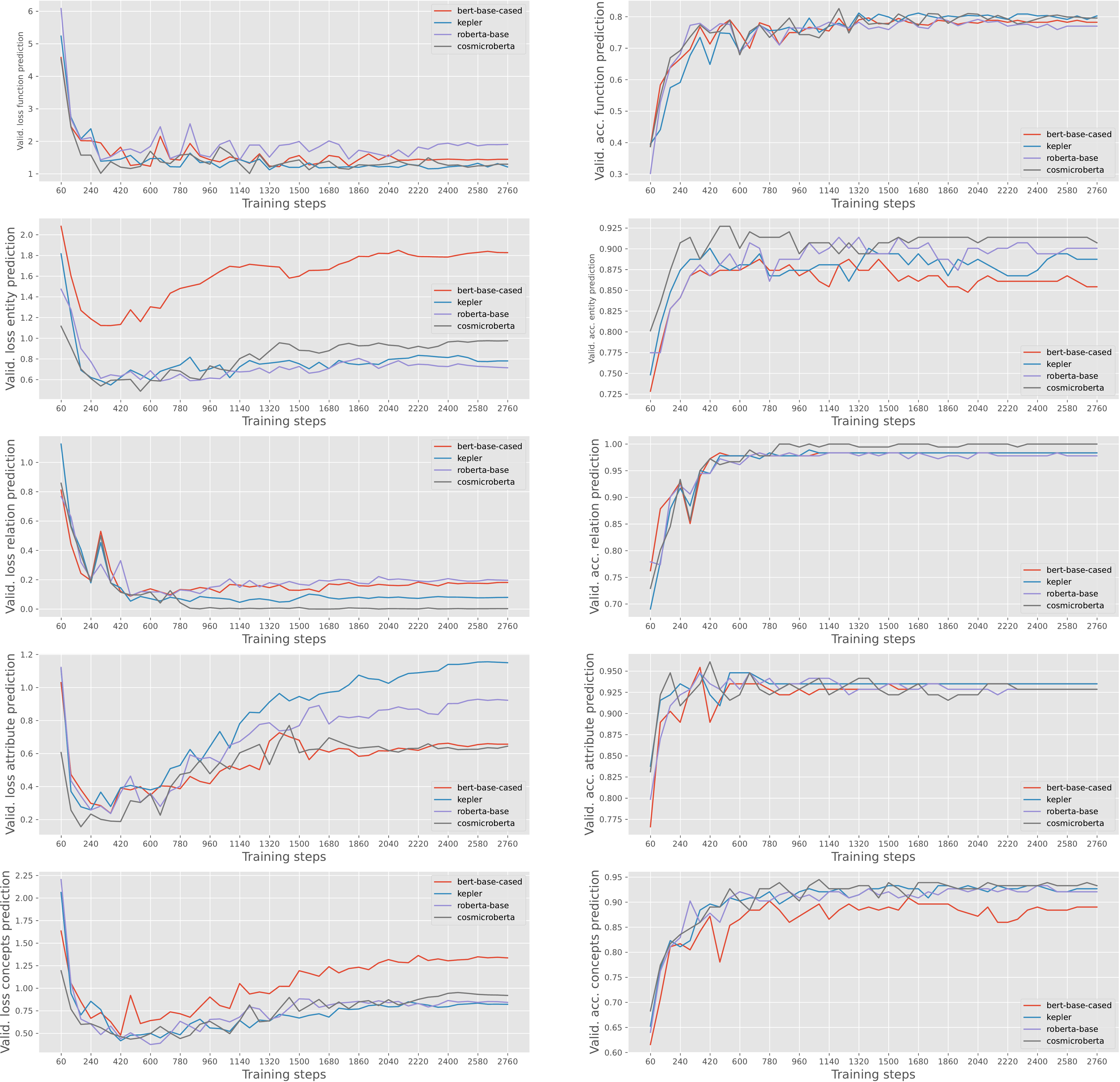}
    \end{center}
    \caption{Validation loss and Accuracy over number of training batches for each component and each tested model.}  
    \label{fig:loss_curves}
\end{figure}
\end{onecolumn}

\begin{twocolumn}

\end{twocolumn}
\end{document}